# Machine-assisted Writing Evaluation:

## Exploring Pre-trained Language Models in Analyzing Argumentative Moves


Wenjuan QIN[1*], Weiran WANG[1], Yuming YANG[2], Tao GUI[23]
College of Foreign Languages and Literature[1]
School of Computer Science[2]
Institute of Modern Languages and Linguistics, Fudan University[3]


## Abstract


The study investigates the efficacy of pre-trained language models (PLMs) in analyzing argumentative moves in a longitudinal learner corpus. Argumentative writing requires a range of complex cognitive skills, including logical reasoning, evidence support, and acknowledgment of opposing positions. Prior studies on argumentative moves often rely on qualitative analysis and manual coding, limiting their efficiency and generalizability. The study aims to: 1) to assess the reliability of PLMs in analyzing argumentative moves; 2) to utilize PLM-generated annotations to illustrate developmental patterns and predict writing quality. A longitudinal corpus of 1643 argumentative texts from 235 English learners in China is collected and annotated into six move types: *claim*, *data*, *counter-claim*, *counter-data*, *rebuttal*, and *non-argument*. The corpus is divided into training, validation, and application sets annotated by human experts and PLMs. We use Bidirectional Encoder Representations from Transformers (BERT) as one of the implementations of PLMs. The results indicate a robust reliability of PLMs in analyzing argumentative moves, with an overall F1 score of 0.743, surpassing existing models in the field. Additionally, PLM-labeled argumentative moves effectively capture developmental patterns and predict writing quality. Over time, students exhibit an increase in the use of data and counter-claims and a decrease in non-argument moves. While low-quality texts are characterized by a predominant use of claims and data supporting only one-side position, mid- and high-quality texts demonstrate an integrative perspective with a higher ratio of counter-claims, counter-data, and rebuttals. This study underscores the transformative potential of integrating artificial intelligence into language education, enhancing the efficiency and accuracy of evaluating students' writing. The successful application of PLMs can catalyze the development of educational technology, promoting a more data-driven and personalized learning environment that supports diverse educational needs.


# 1. Introduction

Argumentative writing demands a multifaceted set of cognitive skills that enable writers to not only support own claims with logical reasoning and evidence but also acknowledge the strengths and weaknesses of opposing positions (Guo et al., 2021; Hsin, 2022). These strategic discourse moves deployed to advance arguments in writing are commonly referred to as *argumentative moves*. Previous research on argumentative move analysis has indicated that proficient writers exhibit a diverse and sophisticated range of moves, such as adept use of evidence to support own claims and explicit acknowledgement of counter-arguments (Fan & Chen, 2021; Toulmin, 2003). However, these studies often rely on qualitative discourse analysis or manual coding of the argumentative structure, limiting the efficiency and generalizability of their findings (O'Halloran, 2011; Strobl et al., 2019; Z. Wang & Chiu, 2023).

The advert of machine learning technology has opened up new avenues for writing evaluation on a large scale. In particular, pre-trained language models (PLMs) represent a type of machine learning model pre-trained on a vast amount of text data (e.g., Google Books, Wikipedia information). PLMs are able to understand not only the meaning of individual words but also the context of words within a sentence by analyzing their relationships with all other words. There are various types of PLMs, including BERT developed by Google and GPT created by OpenAI. In the current study, we use BERT, or Bidirectional Encoder Representations from Transformers,



because it processes each word in the context of all surrounding words, both from the left and the right. In comparison to GPT (which is a left-to-right unidirectional model), BERT excels in understanding contextual meanings and classifying sentences based on their argumentative moves, aligning well with the objectives of the current study (Song et al., 2020; H. Wang et al., 2020; Yamaura et al., 2023).

The current study is among the initial attempts to employ PLMs to analyze argumentative moves utilized in a self-developed English as a Foreign Language (EFL) learner corpus. The study aims to achieve two primary objectives. First, following the widely adopted machine learning protocol, we train the PLM to automatically annotate argumentative moves in EFL learners' writing. The model's reliability and accuracy in argumentative move annotation is evaluated by F1 score, a wide-adopted machine learning matric used to measure how well a model identifies and classifies textual information correctly. Second, on the basis of the validated model, we apply PLM-annotated argumentative moves to address two application inquiries: 1) to distinguish argumentative writing quality levels as evaluated by a group of expert human raters who are blind to the research objectives; 2) to depict individual learners' developmental trajectories over a full academic year. The dual applications of PLM-annotated moves in the study highlight a strategic integration of both text-centered and individual-centered approach in writing evaluation (Qin et al., 2024; Zhou et al., 2023). The first application focuses on analyzing the text itself, identifying argumentative moves predictive of different levels of writing quality as evaluated by human raters. The second approach focuses on individual learners,



tracking how their argumentative moves evolve over time. Modeling individual progress is essential for understanding learning dynamics and offering more personalized feedback and instructional strategies.

## 2. Literature Review

### 2.1. Defining and measuring argumentative moves

In writing research, a *move* is defined as "a discoursal or rhetorical unit that performs a coherent communicative function in a written or spoken discourse" (Swales, 2004, p. 228). A *move* serves to accomplish specific communicative functions, such as an extension of written or spoken discourse. Within the realm of argumentative writing, a discourse aimed at persuading readers of the validity of a central statement, each idea unit in an essay is categorized as incorporating one specific type of *argumentative move* (Reznitskaya et al., 2009; Taylor et al., 2019). The idea unit is segmented by T-unit, or "minimal terminable unit", which is defined as a main independent clause along with any subordinate clauses or non-clausal structures attached to or embedded in it (Hunt, 1970). T-unit is selected as the unit of analysis since it is considered the smallest grammatically allowable unit that can stand alone as a complete sentence.

In the analysis of *argumentative moves*, the present study is guided by two theoretical frameworks: the Toulmin's argumentation model (2003) and Kuhn and Crowell's (2011) developmental model of argumentation. Toulmin's argumentation



model is one of the most widely-adopted model in argumentation studies (Crammond, 1998; Stapleton & Wu, 2015). It categorizes argumentative moves into six types: claim, data, warrant, backing, qualifier, and rebuttal (Toulmin, 2003, pp. 89-94). The argumentative move analysis in the current study (see Section 3.2) is strongly guided by Toulmin's model, which breaks down arguments into several key components to assess their logical structure, including main assertions (*claim*) and their supporting evidence (*data*); as well as moves used to address opposing viewpoints and enhancing the argument's robustness (*counter-claim*, *counter-data*, *rebuttals*). Components such as *warrant* and *backing* are not annotated in the current studry as they typically appear in more advanced argumentative writing produced by expert writers, which rarely appear in learner corpora.

The second theoretical framework, Kuhn and Crowell's (2011) developmental model of argumentation, interprets the evolution of argumentation skills as reflecting changes in learners' evolution of epistemological understanding, which they describing progressing through stages of Absolutism, Multiplism, and Evaluativism (Kuhn et al., 2000). Absolutism views knowledge as objective and decisive; Multiplism recognizes knowledge as diverse, subjective, and indeterminate; Evaluativism, sees knowledge as constructed and in need of evaluation. Learners at this advanced stage are able to check the validity of different perspectives and reach tentative conclusions about certain issues. This developmental process suggests that the sophistication of learners' argumentative writing increases as they advance through these stages, moving from presenting solely their own views to integrating



and evaluating multiple perspectives (Kuhn & Crowell, 2011). Correspondingly, Kuhn and Crowell' developmental model of argumentation encompasses four types of arguments in students' writing, i.e., own side only arguments, dual perspective arguments, integrative perspective arguments, and non-arguments. *Own-side-only arguments* refer to the ideas which only support their own position. *Dual perspective arguments* entail that the ideas offered negatives of opposing positions. *Integrative perspective arguments* refer to the ideas that included a negative about the favored position of a positive about the opposing position. The theoretical assumption underlying the evolution of epistemological understanding perceive that dual perspective and integrative perspective arguments indicate more advanced reasoning (Taylor et al., 2019).

The integration of these two theoretical frameworks enables us to not only analyze argumentative structures using Toulmin's model but also to examine the developmental aspects of these structures as learners' progress over time. Prior studies have shown that as learners enter higher grade levels or become more proficient in the target language, they tend to utilize higher proportion of integrative perspectives in writing (Crowell & Kuhn, 2014; Kuhn et al., 2013). Moreover, the ratio of counter-arguments and rebuttal is found to be significantly predictive of writing quality as evaluated by human raters (Hemberger et al., 2017; Mateos et al., 2018). Our study extends this line of research through the automatic annotation of argumentative moves in a large, longitudinal learner corpus, addressing previous limitations of manual coding and small sample size.



## 2.2. Machine-assisted writing evaluation

In the last two decades, the advert of machine learning, or more specifically natural language processing (NLP) technology, has led to the development of numerous automated tools applicable for writing evaluation. A productive line of research has generated tools for writing evaluation at lexical or syntactic levels. Specifically, machine-assisted lexical analysis focuses solely on information about word tokens and types. A number of lexical analysis tools have been developed and made available, including but not limited to: CLAN (MacWhinney, 2000), VocabProfile (Cobb, 2007), Coh-Metrix (McNamara et al., 2014), LIWC (Pennebaker et al., 2007), LCA (Lu, 2012), and TAALES (Kyle et al., 2018). These tools have been extensively validated in studies focusing on various aspects of language development, such as L2 vocabulary acquisition (Crossley et al., 2010), reading difficulty (Cobb, 2007), and writing quality (Peng et al., 2023). Furthermore, machine-assisted syntactic analysis has seen significant advancements in the recent decade due to the rapid development of NLP techniques in part-of-speech (POS) tagging, constituency parsing, and dependency parsing (Kyle & Crossley, 2018). These techniques involve marking each word in a sentence with a tag that indicates its part of speech (e.g., noun, verb, adjective) and syntactic attributes, such as subject, predicate, and object. This information could be easily processed by computer algorithms, supporting the automatic analysis of syntactic structures. Widely-adopted syntactic parsers include Stanford Parser (Klein & Manning, 2003), Tregex (Levy & Andrew, 2006), and Stanford Neural Network Dependency Parser (D. Chen &



Manning, 2014). These syntactic parsers facilitate the automated creation of constituency representations of sentences. Built upon these, a series of syntactic complexity analyzers were developed, including Biber Tagger (Biber & Conrad, 2014), L2SCA (Lu, 2010; Lu, 2011), and TAASSC (Kyle and Crossley, 2018). Numerous empirical studies have deployed these tools and acknowledged their contribution to the evaluation of writing quality and or proficiency gains (Lei et al., 2023; Qin & Uccelli, 2020; Zhang & Lu, 2022).

Machine-assisted writing evaluation at the discourse level primarily employs three main approaches: cohesion analysis, semantic analysis and summarization based on rhetorical structure theory. To begin with, the Tool for Automatic Analysis of Text Cohesion (TAACO; Crossley et al., 2016; Crossley et al., 2019) exemplifies cohesion analysis by offering a comprehensive analysis of text cohesion, ranging from local cohesion between sentences, global cohesion between chunks of texts, as well as overall text cohesion. Additionally, NLP techniques such as latent semantic analysis (LSA) (Landauer et al., 1998), latent Dirichlet allocation (Blei et al., 2003), and Semantic Web (McClure, 2011) are deployed to examine semantic relations in texts. Finally, various studies have evaluated automatic text summarization based on rhetorical structure theory (Chengcheng, 2010; Hou et al., 2020). Though highly informative and innovative, none of these existing approaches specifically focus on analyze argumentative structure, especially those composed by learner writers. Closest to the current study, AntMover (Anthony, 2003) looked into the argumentative moves of writing, but it merely focused on the introduction part of



academic writing, limiting its generalizability to other genres and full-text analyses. Qualitative studies and manual coding of texts remain the primary choices for argumentative move analysis (Qin & Uccelli, 2021; Strobl et al., 2019; Z. Wang & Chiu, 2023). These gaps highlight a critical shortfall in the ability of current tools to fully capture the complexity of argumentative discourse. Without robust tools to analyze discourse-level elements, evaluation of writing proficiency may overlook critical aspects of text construction that reflect higher-order thinking and linguistic skills. which can result in a partial understanding of a learner's writing development and incomplete writing instruction.

In sum, a review of literature highlights significant advancements in machine-assisted writing evaluation, with substantial progress in lexical and syntactic analyses through NLP-based tools. Despite these developments, notable gaps remain in discourse-level analysis, where current tools lack the capability to fully analyze complex discourse structures and argumentative depth. These limitations underscore a critical need for enhanced NLP tools that can provide more comprehensive evaluations of writing proficiency and support higher-order assessment of writing skills.

### 2.3. The potential of pre-trained language models

The potential of pre-trained language models (PLMs) in analyzing argumentative moves is substantial, heralding a new era in machine-assisted writing evaluation (Sethi & Singh, 2022; Yang et al., 2020). PLMs like BERT and GPT have



demonstrated remarkable proficiency in understanding contextual word meanings and capturing intricate language patterns. PLMs are trained on extensive corpora that include a wide range of linguistic contexts. This training enables them to recognize not only isolated words but also how words are related to each other in different contexts. Such "contextual awareness" is essential for argumentative move analysis, since how statements relate to each other is often what defines their role in an argument. For instance, a claim is usually supported by data, opposed by a counter-claim, and defended by a rebuttal.

In the field of computer science and artificial intelligence, researchers have explored the reliability and applicability of PLMs in analyzing argumentative moves across various types of texts. The related works can be broadly categorized into two types of tasks: 1) argumentative move identification, focusing on the separation of argumentative from non-argumentative text units and identifying boundaries; 2) argumentative move classification, addressing the function of argument moves by classifying identified moves into different types such as claims, data, counter-claims. For instance, for argumentative move identification, Moens et al. (2007) identified argumentative sentences in various text types, achieving 0.738 accuracy. Florou et al. (2013) classified text segments using discourse markers and achieved an F1 score of 0.764. Regarding argumentative move classification, Kwon et al. (2007) proposed consecutive steps for identifying different types of claims, achieving an F1 score of 0.67. Rooney, Wang, and Browne (2012) applied kernel methods for classifying text units, obtaining an accuracy of 0.65. While promising, existing PLMs-based



argumentation analyses reviewed above are mostly validated on polished texts (e.g., news articles or legal documents) or essays written by expert language users. The applicability of these models to learner data remains a significant question. Learner data presents unique challenges to the effectiveness of PLMs such as grammatical errors and unconventional language usage, which are common in texts produced by EFL or ESL learners. These challenges, including incorrect verb tenses, unconventional word choices, and over-simplified or syntactically convoluted sentences, can disrupt the grammatical and semantic consistency that PLMs rely on, leading to potential misinterpretations of sentence meanings as well as their roles in the argumentative structure.

To address these challenges, it is necessary to fine-tune PLMs through training on learner corpora annotated by expert human raters. A high-quality, human-annotated training corpus should exhibit several key characteristics. For example, in the annotation of argumentative moves, each annotation must accurately identify and label the argumentative move according a predefined schema grounded in solid theoretical frameworks. This involves correctly distinguishing between claims, data, counter-claims, rebuttals, non-arguments, etc. Furthermore, annotations must be consistent across different texts and annotators, which ensures that similar argumentative moves are annotated the same way each time. Consistency could be achieved through rigorous training and monitored through inter-annotator reliability tests. Lastly, human annotations should accommodate variations in learner language proficiency to ensure that they are inclusive and representative of the diverse ways



learners construct different types of arguments. This inclusivity means not only identifying obvious argumentative moves but also subtler aspects such as implicit arguments and underlying assumptions. In sum, a high-quality, human-annotated learner corpus can enhance the accuracy and efficiency of fine-turning PLMs to automatically identify and classify argumentative moves.

In conclusion, PLMs have demonstrated substantial potential to revolutionize writing assessment. These models can be fine-tuned to provide detailed, personalized feedback on argumentative writing, helping learners to refine their reasoning and structure. Additionally, by analyzing a large number of texts, PLMs can uncover patterns and common challenges in student writing, offering valuable insights for writing instruction. However, it is important to acknowledge that the limitations of PLMs, particularly in handling learner corpora. Thus, the development of high-quality, human-annotated training corpora is essential to improve the accuracy and efficiency of PLMs in educational applications.

### 2.4. Research questions

The current study seeks to assess the reliability and applicability of pre-trained language models in analyzing argumentative moves utilizing a self-developed, human-annotated learner corpus. The study is guided by two sets of research questions:

**Model validation:**



Research Question 1: To what extent can the pre-trained language model (PLM) effectively and reliably annotate argumentative moves in high school EFL learners' argumentative writing in a manner comparable to human annotations?

**Model application:**

Research Question 2.1: At the textual level, can PLM-annotated argumentative moves distinguish different levels of writing quality, as evaluated by human raters?

Research Question 2.2: At the individual level, can PLM-annotated argumentative moves capture the language development of individual learners over time?

## 3. Methods

### 3.1. The corpus

The study utilized a corpus comprising 1643 argumentative texts composed by 235 EFL learners enrolled in two public high schools in Eastern China. The sample comprised 86 learners in Grade 9 and 149 in Grade 10, ranging in age from 15 to 17 years. Based on a self-reported survey, these learners spoke Mandarin Chinese as their native language and had learned English as a foreign language for approximately 8 to 11 years. Their English proficiency, as assessed by the schools' evaluation report and aligned with the Common Curriculum Standard of the People's Republic of



China (Ministry of Education and State Language Affairs Commission, 2018), ranged from basic (A1) to upper-intermediate (B2) levels.

The corpus was longitudinally compiled over one academic year. During this time period, participants engaged in an educational curriculum designed to enhance high school students' critical thinking and writing ability through discussion and debate. This curriculum employed the Word Generation program, developed by researchers in the United States, known for its efficacy in cultivating adolescents' thinking and language skills in various sociocultural contexts (Jones et al., 2019; Lawrence et al., 2015). Twelve writing topics were carefully selected and adapted to match the language proficiency level and cultural background of the current learner sample. These topics were crated in a way that allow writers to engage in argumentation on socially or scientifically contentious topics. For instance, in writing about the topic "Whether animal testing should be allowed in conducting scientific experiments?", learners were tasked with advocating a position supported by logical reasoning and solid evidence; on the other hand, they were expected to recognize the existence of potential counter-arguments and provide appropriate concessions or rebuttals. Essays were collected on a monthly basis, totaling twelve writing tasks over the academic year.

Prior to data collection, consent was obtained from both learners and their parents, ensuring ethical considerations and adherence to privacy regulations. This project was waived for formal ethics approvals by the Institutional Review Board



(IRB) at the researchers' university for the following reasons. First, the study dealt with existing datasets collected for educational purposes by the school, not specifically for the research project. According to the IRB guideline, secondary data analysis could be exempt from IRB review. Second, the data had been stripped of all identifying information when they were provided to the researchers. There was no way it could be linked back to the subjects from whom it was originally collected, so its subsequent use would not constitute "human subject research".

### 3.2. The argumentative move scheme and human annotation

Leveraging high-quality annotated data to train language models is crucial for enhancing model accuracy and efficiency (Khurana et al., 2023). To achieve this goal, we employed a theory-driven annotation scheme, initially used to train human annotators for reliable data annotation. Subsequently, the annotated data generated by human annotators served as the training set for the language models. Our annotation scheme drew on classic argumentation theories and coding schemes from prior empirical research.

The annotation process began with the segmentation of all texts into T-units, representing independent clauses and their subordinate structures (Hunt, 1970). T-unit was chosen as the primary annotation unit because it represented the smallest thematic units in texts, enabling a fine-grained analysis of argumentative moves. Each T-unit was assigned an argumentative move label based on its contextual meaning and discourse function within the texts.



In our corpus, eight types of argumentative moves were annotated: *title*, *claim*, *data*, *counter-claim*, *counter-data*, *rebuttal-claim*, *rebuttal-data*, *non-arguments*. Figure 1 provided an illustrative example of an annotated argumentative text. The *title* was considered a necessary component in an argumentative text, acting as the hook to attract readers and usually covering the main topic of the essay. *Claims* were assertions in response to a given topic, encompassing both direct responses to the argumentation prompt and additional sub-claims in support of the major claim from different perspectives. *Data* served as evidence supporting a claim, which took various forms such as facts, logical explanations, suppositions, statistics, anecdotes, research studies, expert opinions, definitions, and analogies. *Data* could also include any topic-related non-claim statements, such as introductory sentences introducing background information or rhetorical questions engaging readers. While *claim* and *data* expressed the writer's own opinions, the other four types of moves demonstrated the writer's ability to account for alternative perspectives and provide appropriate responses. Specifically, *counter-claims* presented opposing views challenging the validity of the writer's claims; and *counter-data*, similar to data, served to support counter-claims. *Rebuttal-claims*, usually appearing right after *counter-claims* or *counter-data*, were statements used by the writer to respond to alternative perspectives. *Rebuttal-data* were evidence to support rebuttal claims which included the identification of possible weaknesses in the *counter-claim* or *counter-data*, such as logical fallacies, insufficient support, invalid assumptions, and immoral values.



Finally, *non-arguments* referred to expressions irrelevant to the current topic under discussion or incomprehensible sentences.

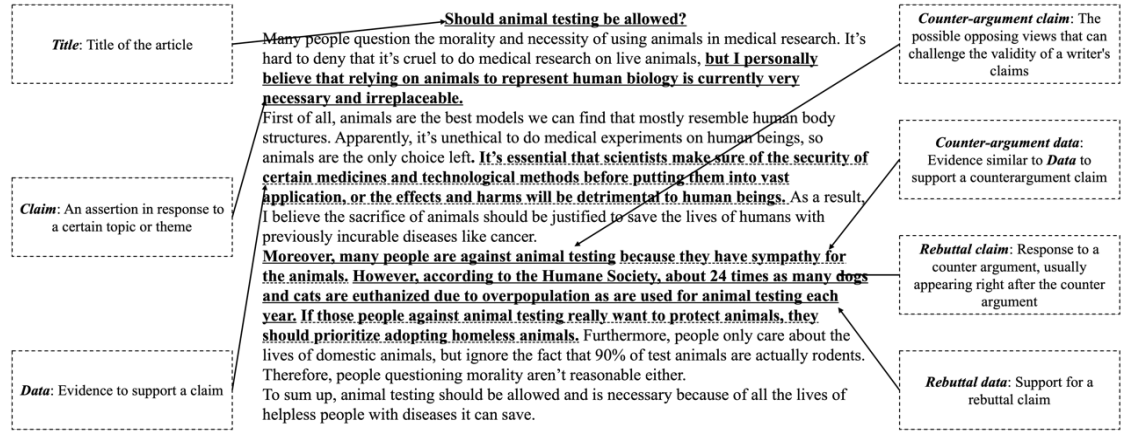

Fig. 1: Example of an annotated argumentative essay

Eight proficient English users, including undergraduate and postgraduate research assistants, were trained to annotate argumentative moves using the scheme. These human experts collaboratively annotated 20% of the training dataset, achieving an ultimate reliability across all annotators ($\kappa = 0.99$). Any disagreements in annotation or confusion about the annotation scheme were addressed during the process of collaborative annotation. Subsequently, the eight human experts independently annotated the remaining 80% of the training dataset. As such, we ensured the annotated data we used for model training had achieved expert consensus and optimal quality.

### 3.3. Model training



Following standard machine learning protocols, we partitioned the corpus into three distinct sets. As shown in Table 1, the *training set* containing 1170 texts (70% of the corpus) was exclusively annotated by human annotators. These annotations were used as ground truth to fine-tune the PLM for our annotation task. The *validation set* containing 234 texts (15% of the corpus) was annotated by both human annotators and PLM, which were used for calculating the F1 score for model validation (see Section 3.5 for F1 calculation algorism). Finally, the *application set* containing 239 texts (15% of the corpus) was only annotated by PLM. The PLM-annotated argumentative moves were then applied to distinguish writing quality levels and predict individuals' writing development. The specific model architecture and training procedure are detailed in the following sections.

Table 1: Corpus composition.

| Research step | Data Set | Corpus size | Annotated by: |
| --- | --- | --- | --- |
| Step 1: Model Validation | Training set | 1170 | human raters only |
| | Validation set | 234 | human raters and PLM |
| Step 2: Model Application | Application set | 239 | PLM only |

### 3.3.1. Problem definition and notations

In the task of argumentative move annotation, we were given an essay, denoted as $d = (w_1, w_2, ..., w_N)$, consisting of $N$ words. The primary objective was to first identify $M$ argumentative moves $(x_1, x_2, ..., x_M)$ by segmenting the words in the essay:



$$d = (\overbrace{w_1, w_2, \ldots}^{x_1}, \overbrace{w_i, \ldots, w_j}^{x_2}, \ldots, \overbrace{w_k, \ldots, w_N}^{x_M})$$

and then classify the identified argumentative moves $(x_1, x_2, \ldots, x_M)$ into specific types, denoted as $(y_1, y_2, \ldots, y_M)$. Here each $y_i \in Y$, where $1 \le i \le M$, represented the type of argumentative move corresponding to $x_1$ and $Y$ was the predefined set of argumentative move types,

$$Y = \{Title, Claim, Data, \ldots\}$$

### 3.3.2. Overall model architecture

An intuitive method would be to use a two-stage approach, addressing the identification and classification tasks separately. However, this method was susceptible to error propagation: inaccuracies in the identification stage could amplify errors in the classification stage. To mitigate this issue, we employed a unified approach that simplified the identification task and could efficiently addresses both tasks within a single classification model. As illustrated in Figure 2, we addressed the argumentative move annotation task in following steps:

### Step 1: **Rule-based candidate argumentative move identification**

For any given essay $d$, we first segmented it into a series of candidate argumentative moves $(x_1^C, x_2^C, \ldots, x_{M_c}^C)$, using hand-crafted rules. Our segmentation rules were primarily based on punctuation marks to identify candidate argumentative moves with the finest granularity. Consequently, we had $M \le M_c$, where $M$ was the



number of actual argumentative moves and $M_c$ was the number of candidate argumentative moves in the essay. Our predefined rules presumed that argumentative moves were delineated by punctuation marks. These rules could also be modified to accommodate edge cases, such as moves separated by "and", when necessary.

Step 2: **Candidate argumentative move classification**

Next, we employed a BERT-based classification model to determine the type of each candidate argumentative move considering contextual information (refer to Section 3.3.3 for a detailed description):

$$y_i^c = CLS(x_i^c, d)$$

where $CLS$ was the trained classification model, $y_i^c$ was the predicted type for $x_i^c$. To address the identification task, we introduced an additional $none$ type to our predefined set of argumentative move types, $Y$. The expanded set of candidate argumentative move types was denoted as $Y^C$:

$$y_i^c \in Y^C, Y^C = Y \cup \{none\}$$

The $none$ type was assigned when a candidate argumentative move should not be individually identified as a correct argumentative move. In this way, we could leverage the same classification model to simultaneously classify and identify argumentative moves, ensuring a more efficient and accurate analysis.

Step 3: **Final argumentative move verification**



Finally, we verified the set of candidate argumentative moves $\{x_i^C\}_{i=1}^{M_C}$ which were initially segmented by rules, based on their predicted candidate types $\{y_i^C\}_{i=1}^{M_C}$. Specifically, we filtered out those candidate argumentative moves classified as $none$ and merged them with adjacent correct candidate moves on the right. This process was encapsulated in the function $MergeInvalid$, which took the candidate moves and their predicted types as inputs and output the final sets of identified and classified argumentative moves:

$$\{x_j\}_{i=1}^{M}, \{y_j\}_{i=1}^{M} = MergeInvalid\left(\{x_i^C\}_{i=1}^{M_C}, \{y_i^C\}_{i=1}^{M_C}\right)$$

The right-first strategy when merging adjacent moves was consistent with our training data preparation procedure described in Section 3.3.3, during which we assigned the ground truth label of a correct move to the right-most candidate move that shared the same ending position.



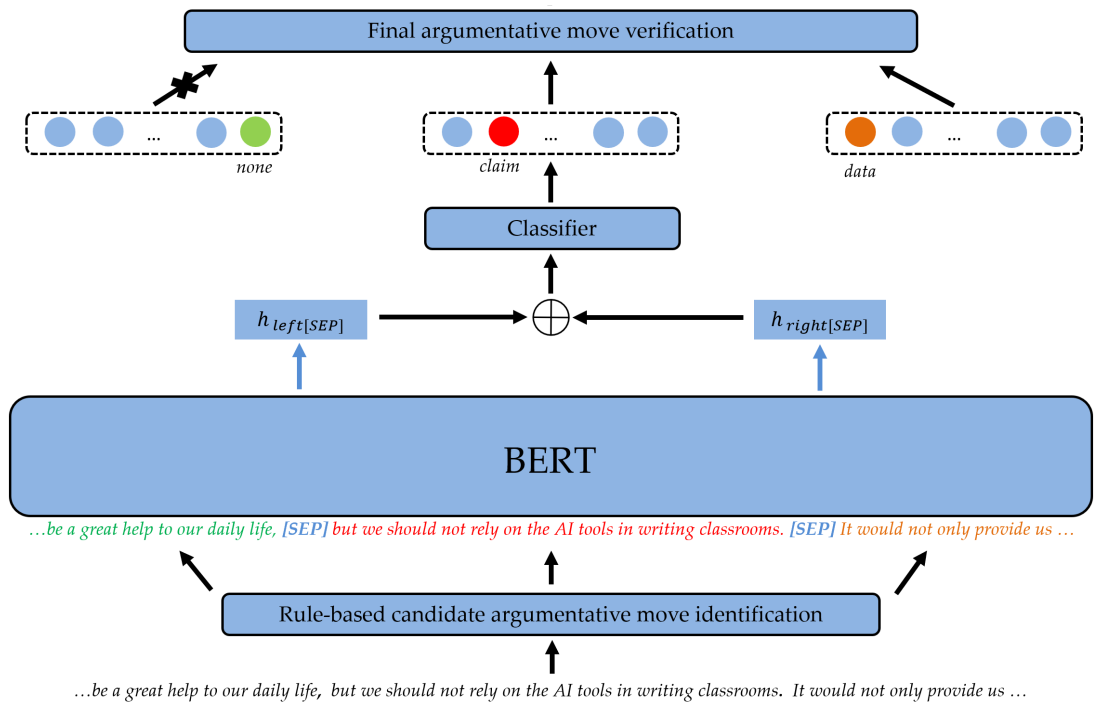

Fig. 2: The architecture of our argumentative move annotation model

### 3.3.3. BERT-based classification model

In this section, we delve into the BERT-based classification model referenced in the previous section. As introduced in Section 2.3, PLMs like BERT (Devlin et al., 2018) can be used as an encoder that analyzes the input text to produce a nuanced, context-aware representation for each word in a sentence. In our model, we added an additional classifier layer on top of BERT encoder and fine-tune BERT for our specific classification task using the annotations in the training corpus. This approach allowed us to harness contextual information effectively, ensuring precise annotation of each candidate argumentative move.



More specifically, our approach began by inserting special $[SEP]$ tokens at the boundaries of candidate argumentative moves within the input essay, forming a modified essay $\tilde{d}$:

$$\bar{d} = concat(x_1^C, [SEP], x_2^c, [SEP], \dots, x_{M_C}^C)$$

The entire essay, now represented as $\tilde{d}$, served as the context and was fed into the BERT encoder. This process yielded output embeddings $\boldsymbol{h}$, capturing the contextual information:

$$\boldsymbol{h} = BERT(\tilde{d})$$

Here, $\boldsymbol{h} \in R^{l \times d}$ where $l$ was the number of tokens in $\tilde{d}$ and $d$ represented the dimension of hidden state.

Next, for each candidate argumentative move $x_i^C$ where $1 \leq i \leq M_C$, we combined the embeddings of the $[SEP]$ tokens on its left and right sides and used this combined embedding as input for the classifier. The classifier then output the predicted candidate argumentative move type $y_i^C$:

$$y_i^C = f_{classifier}(\boldsymbol{h}_{left[SEP]ofx_i^C} + \boldsymbol{h}_{right[SEP]ofx_i^C})$$

Here, $y_i^C$ represented the most probable type from the set $Y^C$ as determined by the classifier for the candidate argumentative move $x_i^C$.

For each essay in the training corpus, we knew the actual argumentative moves and their corresponding ground truth annotations, represented as actual annotation



pairs $(x_j, y_j), 1 \leq j \leq M$. To train our model, it was essential to establish ground truth annotations for candidate argumentative moves, denoted as candidate annotation pairs $(x_i^C, y_i^C), 1 \leq i \leq M_C$, using these actual pairs. This was achieved by matching each candidate move $x_i^C$ with an actual move $x_j$ based on their shared ending position. Consequently, the ground truth type for $x_i^C$ was assigned as $y_i^C = none$, indicating an invalid argumentative move split.

We utilized the pre-trained BERT-base-uncased model as our BERT encoder. The classifier consisted of a linear layer followed by a softmax function. Both the BERT encoder and the classifier were fine-tuned during training, utilizing cross-entropy loss to optimize their parameters.

*3.4. Human scoring of writing quality*

To assess the holistic writing quality of argumentative essays, we employed a scoring rubric validated in prior studies (Deng et al., 2022; Phillips Galloway et al., 2020). Informed by the NAEP (2011) Writing Framework, this rubric comprehensively evaluated writing quality across four dimensions: The *Position* dimension assessed the number of perspectives taken into consideration. The *Organization* dimension evaluated the logical and coherent structuring of the essay. The *Support* dimension measured the degree of depth, complexity, elaboration, and connectedness of ideas provided in texts in support of the position. The *Clarity* dimension gauged the precision and unambiguous conveyance of information in an essay (see Appendix 1).



Each dimension was scored on a four-point scale, with higher scores indicating superior quality. Then these dimensional scores were holistically considered to determine the final designation of writing quality levels. Specifically, high-level writing comprised essays scored at 4 points on at least three dimensions; medium-level writing included those scored at 3 or 4 on at least two dimensions; and low-level writing included those scored at 2 or below on all dimensions.

A team of four human raters, blinded to the research objectives, received rigorous training for scoring consistency. To ensure inter-rater reliability, we used 20% of the data for double-scoring and achieved $\kappa = 0.89$. Then, four raters independently scored the remaining 80% of all data. This robust scoring rubric and trained raters contribute to the reliable and objective evaluation of the holistic writing quality in argumentative essays.

*3.5. Data analysis*

To address the model validation question, we evaluate the F1 score of the holistic model and that for each type of argumentative move. The F1 score is a widely used metric in classification tasks, particularly in the context of annotation. The F1 score is calculated as follows:

$$F1 = 2 \times \frac{Precision \times Recall}{Precision + Recall}$$

The F1 score is a harmonic mean of precision and recall. Precision is the ratio of true positive predictions to the total positive predictions made. It is formulated as:



$Precision = \frac{True\ Positives}{True\ Positives + False\ Positives}$. Recall is the ratio of true positive

predictions to the total actual positive instances. It is expressed as: $Recall =$

$\frac{True\ Positives}{True\ Positives + False\ Positives}$. A high F1 score indicates that the model has a robust

balance of precision and recall. This is crucial in annotation tasks where both the

avoidance of false positives and the inclusion of all relevant instances are important.

To address the first application question, we conducted one-way between-groups

multivariate analysis of variance (MANOVA). This analysis assessed the overall

efficacy of PLM-annotated argumentative moves on differentiating writing quality

levels as assessed by human raters. Post hoc one-way ANOVA with Bonferroni

correction was employed to pinpoint the most influential argumentative moves and

their effect sizes.

For the second application question, we conducted multi-level regression

analysis, recognizing the nested data structure of argumentative essays within

individuals in the longitudinal corpus. In a series of multi-level regression models,

different types of PLM-annotated argumentative moves were used as the outcome

variables and time served as the individual-level predictor. This approach

incorporated the random variation of intercepts across individuals, thereby capturing

both within-individual and between-individual variability in the models. The

significance of the time coefficient indicated the degree to which individuals adjust

their usage of argumentative moves over time.

## 4. Results



*4.1. Evaluating model reliability*

Overall, the pre-trained language model achieves high reliability in the argumentative move annotation task, with a holistic F1 score of 0.743. This high F1 score indicates that the model has a robust balance of precision (0.73) and recall (0.74). This is crucial in labeling tasks where both the avoidance of false positives and the inclusion of all relevant instances are important.

However, the PLM demonstrates different degrees of accuracy in annotating specific types of argumentative moves (see Table 2). The F1 scores for three types of moves are generally high, with an F1 score of 0.73 for *claims*, 0.76 for *data*, and 0.96 for *title*. The label *none* also has a high F1 score of 0.88, indicating high accuracy in argumentative move segmentation. In addition, the F1 scores for three types of argumentative moves have achieved medium accuracy, with an F1 score of 0.42 for *counter-claims* and *rebuttal-claims*, and 0.46 for *non-arguments*. However, the reliability for two types of moves is not ideal, with the F1 scores lower than 0.30 for *counter-data* and *rebuttal-data*. The low F1 scores might be explained by the limited sample in the training data. For instance, among the 1170 texts annotated by humans, only 121 *counter-data* and 136 *reb-data* are identified. The limited training set fails to provide sufficient ground truth for model training, resulting in lower reliability for model annotation. Due to the low reliability, these two types of argumentative moves are not used in subsequent application analyses.



Table 2: The distribution of PLM-annotated argumentative moves.

| Annotation | Precision | Recall | F1 | Cases |
|---|---|---|---|---|
| [claim] | 0.75 | 0.72 | 0.73 | 1168 |
| [data] | 0.73 | 0.79 | 0.76 | 1880 |
| [title] | 0.93 | 0.99 | 0.96 | 68 |
| [ctarg-claim] | 0.47 | 0.37 | 0.42 | 202 |
| [ctarg-data] | 0.29 | 0.13 | 0.18 | 121 |
| [reb-claim] | 0.51 | 0.36 | 0.42 | 144 |
| [reb-data] | 0.38 | 0.22 | 0.28 | 136 |
| [non-arg] | 0.48 | 0.44 | 0.46 | 32 |
| [none] | 0.84 | 0.92 | 0.88 | 1147 |
| Total | 0.73 | 0.74 | 0.74 | 4898 |

*4.2. Application 1: Distinguishing writing quality*

Our first application inquiry is whether PLM-annotated argumentative moves could effectively distinguish writing quality levels as assessed by human raters. One-way between-group multivariate analysis of variance (MANOVA) reveals that PLM-annotated argumentative moves can significantly differentiate low-, medium- and high-level writing quality ($F_{(14, 820)}$=1.59, $p \leq 001$, Wilk's Lambda=0.95). Further investigation through post hoc one-way analysis of variance (ANOVA) unveils specific types of argumentative moves contributing significantly to this differentiation. Notably, the ratio of *counter-claims* emerge particularly discriminative. High-quality essays demonstrate a distinctive pattern, containing an average of 9% *counter-claims* on average, while medium- and low-quality essays exhibit a lower ratio of such moves, averaging at 5% and 4%, respectively. This cross-level distinction is statistically significant (F=5.71, $p \leq 001$, $\eta^2$=0.027). The other four types of argumentative moves, though capturing interesting descriptive



patterns of cross-level variations, do not demonstrate statistically significant results after Bonferonni correction for multiple comparisons. These findings underscore the nuanced nature of argumentative moves in distinguishing human-rated writing quality, with a particular emphasis on the prevalence of *counter-claims* in high-quality texts. Such insights contribute to a more comprehensive understanding of the intricate components influencing the assessment of argumentative writing quality by human raters.

Table 3: MANOVA analysis results of using PLM-annotated argumentative moves to differentiate writing quality.

|  | Low | Medium | High | Between-group Comparison | |
|---|---|---|---|---|---|
|  | Mean (SD) | Mean (SD) | Mean (SD) | F | $\eta^2$ |
| Claim | 0.35 (0.15) | 0.32 (0.13) | 0.33 (0.13) | 1.13 | 0.005 |
| Data | 0.54 (0.18) | 0.57 (0.19) | 0.53 (0.16) | 0.27 | 0.001 |
| Counter-claim | 0.05 (0.08) | 0.06 (0.04) | 0.09 (0.05) | **5.71\*\*\*** | **0.027** |
| Rebuttal-claim | 0.05 (0.07) | 0.05 (0.06) | 0.06 (0.06) | 1.28 | 0.006 |
| Non-argument | 0.01 (0.02) | 0.01 (0.02) | 0.01 (0.02) | 0.78 | 0.004 |

MANOVA $F(14, 820)=1.59$, $p<.001$, Wilk's Lambda=0.95

*Note.* $*p < .05$, $**p < .01$, $***p < .001$

### 4.3. Application 2: Portraying individual development

The second application inquiry seeks to explore the predictive capacity of PLM-annotated argumentative moves on individual development over time. Through the application of multi-level regression analysis, three specific types of PLM-annotated argumentative moves - i.e., *data*, *counter-claims*, *non-arguments* - emerge as significant contributors to language development across time intervals (see Table 4).



Table 4: Multi-level regression models using PLM-annotated argumentative moves to predict language development over time.

|                    | Claim        | Data         | Counter-claim | Rebuttal-claim | Non-argument |
|--------------------|--------------|--------------|---------------|----------------|--------------|
| *Fixed effects*    |              |              |               |                |              |
| wave               | 0.02         | **1.28**\*\*\* | **0.17**\*   | 0.04           | **-0.11**\*  |
| Intercept          | 33.75\*\*\*  | 51.65\*\*\*  | 4.20\*\*\*    | 2.61\*\*\*     | 1.02\*\*\*   |
| *Random effects*   |              |              |               |                |              |
| Between-individual | 1.70\*\*\*   | 1.87\*\*\*   | -0.49         | 0.59\*\*\*     | -0.16        |
| Within-individual  | 2.51\*\*\*   | 2.77\*\*\*   | 1.80\*\*\*    | 1.34\*\*\*     | 0.58\*\*\*   |
| AIC                | 3781.67      | 4015.77      | 3042.16       | 2689.98        | 1974.16      |
| BIC                | 3798.29      | 4032.39      | 3058.78       | 2706.60        | 1990.78      |

*Note.* \**p* < .05, \*\**p* < .01, \*\*\**p* < .001

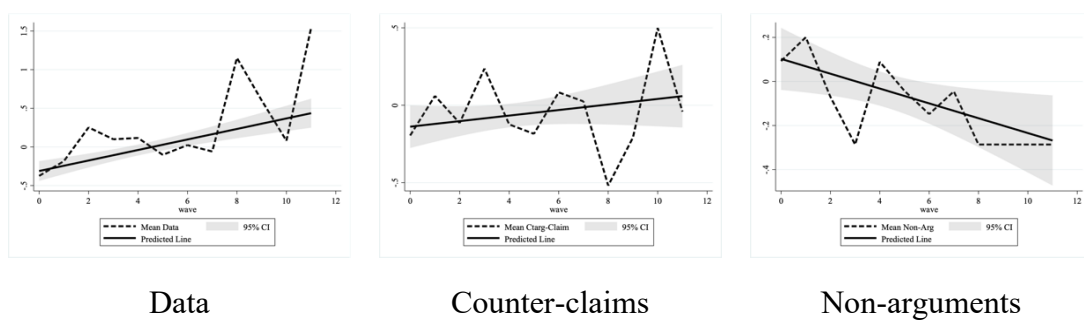

| Data | Counter-claims | Non-arguments |

Fig. 3: Developmental trajectories of *data*, *counter-claims*, and *non-arguments*.

As illustrated in Figure 3, at the individual level, the predictor time exhibit noteworthy associations argumentative moves. Specifically, positive correlations are identified between time and the ratio of *data* (β = 1.28, *z* = 4.63, *p*<.001) and *counter-claims* (β = 0.17, *z* = 2.12, *p* = 0.04), indicating that over one academic year, EFL learners tend to increase their utilization of argumentative moves supporting own claims with concrete evidence and acknowledging counter-arguments. Conversely, a negative association is found between time and the ratio of *non-arguments* (β = −0.11, *z* = −3.42, *p* = 001), suggesting a significant decrease in the proportion of irrelevant or



incomprehensible expressions. These findings underscore the predictive capacity of PLM-annotated argumentative moves in capturing individual EFL learners' developmental trajectory. The observed shifts in the use of *data*, *counter-claims*, and *non-arguments* reflect nuanced changes in language proficiency over the full academic year. Notably, associations between time and other types of argumentative moves were non-significant, emphasizing the specificity in the development of argumentative skills.

## 5. Discussion and Implication

### 5.1. Opportunities and challenges of using PLMs for writing evaluation

The first objective of our study is to evaluate the reliability of the pre-trained language model (PLM) in annotating argumentative moves in high school EFL learners' writing. The overall performance of the PLM in this task is promising, as indicated by a high F1 score of 0.743. Our findings align with the broader landscape of PLM applications in argumentative move identification and classification. Existing studies (Devlin et al., 2018; Mikolov et al., 2013) have demonstrated the potential of PLMs, such as BERT and GPT, in understanding contextual word embeddings and capturing intricate language patterns. Contextual awareness enables a nuanced understanding of relationships between statements, facilitating the identification of claims, data, counter-claims, rebuttals, and non-arguments in argumentative writing.



However, upon closer examination, the PLM demonstrates varying degrees of accuracy in annotating specific types of argumentative moves. Notably, three move types, *claims*, *data*, and *title*, exhibited high F1 scores. The label *none* also achieved a high F1 score of 0.88, indicating accurate segmentation of argumentative moves. In contrast, for *counter-claims* and *rebuttal claims*, as well as *non-arguments*, the PLM demonstrated only moderate accuracy. In contrast, the reliability for *counter-data* and *rebuttal-data* was suboptimal, because of the limited representation of these move types in the training data. Such finding highlights the needs for continuous model improvement targeting language learner corpora. Our study points to several implications worthy of further investigation. First, learner data often encompasses a wide range of language proficiency levels, with variations in grammatical accuracy and language conventions. PLMs, while proficient in understanding contextual word meanings, may encounter challenges when faced with grammatically erroneous or unconventional language use (Qu & Wu, 2023; H. Wang et al., 2023). Consider the following example from the corpus where a student discusses the topic, "Does rap music have a negative impact on youth?". The annotations were initially made by PLM and later corrected by human experts:

> *In conclusion, listening to rap is just as listening to other kinds of music* [claim]*,*
> *__and__ we should choose the extents of rap carefully __and__ avoid being influenced by*
> *the negative concepts in a rap* [PLM: data]/[Human: counter-claim]*.*



In this statement, the learner aims to present a balanced perspective on the influence of rap music on youth. However, the incorrect use of the conjunction "and" leads BERT to mislabel the second part of the sentence as [data]. Upon closer examination, human experts classify it as [counter-claim], recognizing the acknowledgement of rap music's potential negative impacts and suggesting a cautious approach. A more suitable discourse marker, such as "but," to signal the adversative relationship between the two T-units might have enabled PLMs to label the sentence more accurately.

Second, the limited representation of certain move types in our training data highlights the necessity for specialized annotation efforts when dealing with learner data. Annotated corpora specifically designed to capture the intricacies of argumentative moves in learners' writing can enhance the PLM's ability to handle diverse linguistic expressions, including the recognition of non-standard language structures, common grammatical errors, and typical rhetorical patterns in learner writing. However, these efforts are resource-intensive, requiring significant time, expertise, and financial investment. There is also a risk of introducing biases based on annotators' interpretations, which could skew the training of PLMs. Moreover, annotated corpora can quickly become outdated, necessitating ongoing updates to remain relevant and effective for training purposes.

Third, PLMs trained on general text corpora might face challenges in generalizing across different proficiency levels present in learner data. Fine-tuning



PLMs with learner-specific data can contribute to better model performance. For instance, researchers could select representative writing samples from a broad spectrum of proficiency levels and diverse educational settings, curating a balanced dataset that includes beginner, intermediate, and advanced learner texts. These samples should include error-prone language, which is essential for training PLMs to not only recognize correct language structures but also adapt to linguistic variations. Finally, recognizing the dynamic nature of language development in learners, continuous model improvement is essential. Regular updates and refinements to PLMs based on feedback from annotated learner corpora can enhance their adaptability and efficacy in assessing argumentative writing across various proficiency levels.

*5.2. Pedagogical implications of machine-assisted writing evaluation*

The application of the pre-trained language model (PLM) in our study proves instrumental in both distinguishing writing quality levels and predicting individual development over time. Our findings highlight the significant capacity of PLM-annotated argumentative moves to differentiate low-, medium-, and high-level writing quality, providing a nuanced understanding of automated analysis in categorizing writing proficiency. Further investigations unveil specific types of argumentative moves (i.e., *counter-claims*) contributing significantly to this differentiation. In the realm of predicting individual language development, our exploration identifies specific PLM-annotated argumentative moves — *data*, *counter-claims*, and *non-*



*arguments* — as significant contributors across time intervals. These findings collectively underscore the potential of PLMs not only in assessing writing quality but also in offering predictive insights into individual language development.

These findings hold several pedagogical implications for educators in EFL writing classrooms. By leveraging the discriminative power of specific move types, educators can design targeted instructional interventions. For instance, the emphasis on *counter-claims* suggests a need for explicit instruction in counter- argumentation skills, while the predictive power of *data* indicates the importance of fostering skills related to integrating supporting evidence (Kuhn et al., 2013; Stapleton & Wu, 2015). This pedagogical approach aligns with the fundamental principles of argumentative writing, which posits that a well-developed argument should not only present the main thesis (*claims* and *data*) but also anticipate and address opposing viewpoints (*counter-claims*). This reflects elements of the Toulmin Model of Argumentation (Toulmin, 2003) as well as the epistemological understanding of world knowledge (Chen, 2019), where counter-arguments contribute to the robustness and persuasiveness of an argument. Educators can design explicit lessons and activities that guide students in recognizing, formulating, and effectively incorporating *counter-claims* in their argumentative writing. This might include analyzing examples of effective counter-arguments in texts, engaging in class discussions on opposing viewpoints, and practicing the incorporation of counter-claims in their own writing. For example, consider a student who writes a one-sided claim: "*Cloning should be forbidden as it is immoral.*" Teachers can facilitate deeper analysis and critical



thinking by posing follow-up questions that encourage the addition of supporting data for the claim. They might ask, "What specific ethical principles does cloning violate?" Additionally, to promote more balanced arguments, teachers could prompt students to explore the topic from multiple perspectives by asking, "What are some potential positive aspects of cloning?" This approach guides students towards a more nuanced understanding of the controversial issue, enabling them to construct well-rounded arguments.

## 5.3. Limitations

While our study contributes valuable insights into the reliability and applicability of pre-trained language models (PLMs) in analyzing argumentative moves in high school EFL learners' writing, it is essential to acknowledge certain limitations. Firstly, our investigation primarily focuses on a specific set of argumentative moves, and the generalizability of our findings to other genres or writing tasks may be limited. Additionally, the reliance on a particular PLM architecture may influence the outcomes, and the rapidly evolving landscape of language models may introduce new developments or models that are not considered in this study. Furthermore, the limited size of the training data, particularly for certain argumentative move types, may have impacted the PLM's performance and generalization. The study's context, involving high school EFL learners, may limit the broader application of our findings to different age groups or proficiency levels. Finally, while we explore the predictive capacity of PLM-annotated moves on language development, external factors



influencing learners' progress are not exhaustively considered. Recognizing these limitations provides a foundation for future research to build upon and refine our understanding of PLMs in the context of language learning and writing assessment.

## 6. Conclusion

In conclusion, our study illuminates the potential and challenges associated with integrating pre-trained language models (PLMs) into the evaluation of argumentative moves in high school EFL learners' writing. The high reliability of the PLM in annotating argumentative moves, coupled with its capacity to distinguish writing quality and track individual language development, underscores its utility as a valuable tool in the educational landscape. The discriminative power of specific move types, such as *counter-claims* and the predictive importance of *data*, provide actionable insights for educators seeking to enhance writing instruction. However, acknowledging the study's limitations, including the specificity of the argumentative moves examined and the contextual constraints of the learner population, emphasizes the need for careful consideration in generalizing findings and further development for human-annotated learner corpora. As the field of PLMs continues to evolve, our study contributes a foundational understanding that paves the way for future research, offering opportunities to refine methodologies and explore the broader implications of incorporating PLMs in language learning and writing assessment.